\documentclass{article}
\usepackage{spconf,amsmath,graphicx}
\usepackage{txfonts}
\usepackage{graphicx}
\usepackage{amsmath,amssymb}
\usepackage{subcaption}
\usepackage[square, comma, sort&compress, numbers]{natbib}
\usepackage{url}

\title{EFFICIENT IMAGE CATEGORIZATION WITH SPARSE FISHER VECTOR}
\name{Xiankai Lu$^{1}$, Zheng Fang$^{2}$, Tao Xu$^{2}$, Haiting Zhang$^{3}$, Hongya Tuo$^{2}$\thanks{*Corresponding author: carrierlxk@gmail.com}}
\address{ $^{1}$Department  of Automation, Shanghai Jiao Tong University,Shanghai, China \\
          $^{2}$School  of Aeronautics and Astronautics, Shanghai Jiao Tong University, Shanghai, China  \\
          $^{3}$School of Control Science and Engineering, Shan Dong University, Jinan, China}
\begin{document}
\ninept
\bibliographystyle{unsrt}
\maketitle
\begin{abstract}
In object recognition, Fisher vector (FV) representation is one of the state-of-art image representations ways at the expense of dense, high  dimensional features and increased computation time. A simplification of FV is attractive, so we propose Sparse Fisher vector (SFV). By incorporating locality strategy, we can accelerate the Fisher coding step in image categorization which is implemented from a collective of local descriptors. Combining with pooling step, we explore the relationship between coding step and pooling step to give a theoretical explanation about SFV. Experiments on benchmark datasets have shown that SFV leads to a speedup of several-fold of magnitude compares with FV, while maintaining the categorization performance. In addition, we demonstrate how SFV preserves the consistence in representation of similar local features.
\end{abstract}
\section{Introduction}
\label{sec:intro}

The Fisher vector approach ~\cite{DBLP:conf/cvpr/PerronninD07,DBLP:conf/eccv/PerronninSM10,DBLP:journals/ijcv/SanchezPMV13,DBLP:conf/cvpr/JegouDSP10,DBLP:conf/cvpr/PerronninLSP10,DBLP:conf/iccv/CinbisVS13} is an extension of the popular Bag-Of-Words (BOW) model by encoding for codeword the mean and variance of local descriptors. It consists of two steps (i) encoding step, encoding the descriptors into dense and high-dimensional features codes; and (ii) pooling step, pooling the codes into a vector. With several improvements ~\cite{DBLP:conf/eccv/PerronninSM10,DBLP:journals/ijcv/SanchezPMV13}, Fisher vector has been one of the most effective ways for image categorization.

The success of FV representation is ascribed to its high dimensionality, but FV representation also suffers high computation cost when compared to BOW model~\cite{DBLP:journals/ijcv/SanchezPMV13}, especially for large-scale image retrieval and object detection. For specific tasks, several simplified ~\cite{DBLP:conf/cvpr/JegouDSP10,DBLP:conf/cvpr/OneataVS14,DBLP:conf/cvpr/SandeSS14} and extended~\cite{DBLP:conf/eccv/PerronninSM10,DBLP:conf/cvpr/PerronninLSP10,DBLP:conf/iccv/CinbisVS13,DBLP:conf/icassp/LinDHG13} versions of Fisher coding have emerged. In ~\cite{DBLP:conf/cvpr/JegouDSP10}, Jegou et al. proposed the VLAD representation in which each local descriptor is assigned to the nearest visual word, then the differences between codewords and corresponding descriptor are accumulated. In ~\cite{DBLP:conf/cvpr/PerronninLSP10}, Florent Perronnin et al. compressed the high-dimensional Fisher vectors through Local Sensitive Hashing. Recently, Dan Oneata et al.~\cite{DBLP:conf/cvpr/OneataVS14} presented approximations to normalizations in Fisher vector. In ~\cite{DBLP:conf/cvpr/SandeSS14}, authors realized a fast local area independent representation by representing the picture as sparse integral images. In this paper, we combine the locality strategy into Fisher vector to reduce the time consumption in feature coding step.

”Locality” strategy has been used in Linear Embedding and spectral clustering i.e. Local Linear Embedding~\cite{Roweis2000} and  ‘local’ spectral clustering ~\cite{DBLP:conf/nips/NgJW01}. Inspired by this strategy, many localized coding ways or nearest search algorithms have emerged in BOW, for example, Locality-constrained Linear Coding (LLC)~\cite{DBLP:conf/cvpr/WangYYLHG10}, Local Soft Coding (LSC)~\cite{DBLP:conf/iccv/LiuWL11}, Laplacain Sparse Coding~\cite{DBLP:conf/cvpr/GaoTCZ10}, Local Coordinate Coding ~\cite{DBLP:conf/icml/YuZ10}, local sparse coding ~\cite{DBLP:conf/eccv/YangYH10}. Also this locality-preserving method has been used in pooling step~\cite{DBLP:conf/iccv/BoureauRBPL11}. This locality will produce an “early cut off” effect to remove the unreliable longer distances. Previous work has shown the effectiveness of preserving configuration space locality during coding, so that similar inputs lead to similar codes~\cite{DBLP:conf/cvpr/GaoTCZ10}.
Also we can view them as a trick whose computational cost would be prohibitive with standard coding. Because all the coding coefficients can be regarded as the probability density to describe the feature which can be represented by histogram or fisher vector. In this paper, we will introduce the LLC, LSC and SFV from probabilistic perspective and reveal the relationships between LLC, LSC and SFV, and this part will be discussed in section 4.

For the pooling step in image categorization, Naila Murray~\cite{DBLP:conf/cvpr/MurrayP14} et al. tried to generalize max pooling (GMP) to Fisher vector by constructing object function with loss term. Based on this structure, we reformulate the sparse Fisher vector which is the origin Fisher vector combined with locality strategy.

A notable previous idea which is similar to our work is proposed in ~\cite{DBLP:journals/ijcv/SanchezPMV13} with "posterior thresholding". But ~\cite{DBLP:journals/ijcv/SanchezPMV13} only regarded this as an accelerating trick, and failed to provide the detailed theoretical proof and the effectiveness of the proposed method are not explained. Our paper provides the detailed explanation of the scheme and implement a experimental evaluation on image categorization task.

\section{Generalized max pooling revisit}
\label{sec:format}
Fisher vector is essentially the sum pooling of encoded SIFT features. It should be noted that the sum-pooled representation is more influenced by frequent descriptors in one image. While max-pooled representation only considers the greatest response, and therefore immune to this effect, but it does not apply to aggregation-based encoding such as FV representation. To alleviate the problem, ~\cite{DBLP:conf/cvpr/MurrayP14} proposed the generalized max pooling method that mimics the desirable properties of max pooling. They denote $\phi_n$ the code vector of each feature, and $\phi^{max}$ the GMP vector. GMP demands that $\phi _{\rm{n}}^{\rm{T}}{\phi ^{{\rm{max}}}} = {\rm{Const}}$, which indicates that $\phi^{max}$ is equally similar to frequent and rare features. In the BOF case, GMP is strictly equivalent to max pooling~\cite{DBLP:conf/cvpr/MurrayP14}. 
GMP can be formalized in two ways. The first is the primal formulation:
\begin{equation}
\phi ^{gmp} = \mathop {\arg \min }\limits_\phi  {\left\| {{\Phi ^T}\phi  - {\textbf{1}_{\rm{N}}}} \right\|^2}
\end{equation}
	 	which directly gives the result of pooling $\phi^{gmp}$, where $\textbf{1}_{\rm{N}}$ is the $\emph{N}$-dimensional vector of all ones. The second is the dual formulation:	
\begin{equation}
\alpha _\lambda  = \mathop {\arg \min }\limits_\alpha  {\left\| {{\Phi ^T}\Phi \alpha  - {\textbf{1}_{\rm{N}}}} \right\|^2} + \lambda {\left\| {\Phi \alpha } \right\|^2}\
\end{equation}
which gives the weight of each feature. $\phi^{gmp}$ is the result of weighted sum pooling.

\section{SPARSE FISHER VECTOR THEORY}
\label{sec:pagestyle}
Let $X = \{ {x_1},....{x_N}\} $ be a set of $\emph{N}$ local descriptors extracted from an image. We denote $\emph{M}$ the number of Gaussian Mixture Model(GMM) clusters, and $\emph{D}$ the dimension of SIFT descriptors after using PCA. Clearly.
According to Section 2, the Fisher vector representation $\phi$ should be equally similar to each Fisher vector code, which is defined as:
\begin{equation}
{\Phi}^{{T}}\phi  = {\textbf{1}_{{N}}}\
\label{eq:primal}
\end{equation}
where $\Phi$ is the code matrix, of which each row represents a Fisher vector code corresponding to the descriptor  $\emph{x}_{n}$.
\begin{equation}
{\Phi} = \left( {\begin{array}{*{20}{c}}
{{{{G}}_1}\left( {{{{x}}_1}} \right)}& \cdots &{{{{G}}_{{M}}}\left( {{{{x}}_1}} \right)}\\
 \vdots & \ddots & \vdots \\
{{{{G}}_1}\left( {{{{x}}_{{N}}}} \right)}& \cdots &{{{{G}}_{{M}}}\left( {{{{x}}_{{N}}}} \right)}
\end{array}} \right)
\end{equation}
where $G_m (x_n)$ is the sub-vector of cluster $\emph{m}$ of the Fisher vector code corresponding to ${\emph{x}_{n}} $.

In ~\cite{DBLP:conf/cvpr/PerronninD07}, the normalization of the Fisher information matrix takes a diagonal form, which assumes the sub-vectors are independent of each other. Therefore it is natural to divide Eq. \ref{eq:primal} into $\emph{M}$ subtasks. We denote by $\Phi_m$ the $\emph{m}$-th column of $\Phi$, which is the code matrix in the $m$-th subtask:
\begin{equation}
{{{\Phi }}_{{m}}} = \left( {\begin{array}{*{20}{c}}
{{{{G}}_m}\left( {{{{x}}_1}} \right)}\\
 \vdots \\
{{{{G}}_m}\left( {{{{x}}_{{N}}}} \right)}
\end{array}} \right)
\end{equation}
And $\Phi=(\Phi_{1}\cdots\Phi_{M})$. If each subtask is fulfilled as follows, the whole task likes Eq. \ref{eq:primal} will be fulfilled as:
\begin{equation}
\Phi^T_m \phi_m = \textbf{1}_{{N}}
\end{equation}

The objective function of the $m$-th subtask in the primal formulation is:
\begin{equation}
\phi_{m, gmp} = \arg\min \left \| \Phi^T_m \phi_m - \textbf{1}_{{N}} \right \| + \lambda \left \| \phi_m \right \|^2_2
\end{equation}
Clearly the primal formulation does not have the sparsifying effect, so we turn to the dual formulation. According to Section 2, we denote by $\alpha_m$ the code weight so that $\Phi_m\alpha_m=\phi_m$, which means that $\phi_m$  is  the pooling result of code matrix $\Phi_m$ with weight $\alpha_m$ ~\cite{DBLP:conf/iccv/BoureauRBPL11}. $\alpha_m$ is consistent with the idea of Sparse Fisher vector because it can determine whether a Fisher vector code is valid in the final image representation.

The objective function of the $m$-th task in the dual formulation is:
\begin{equation}
{{{\alpha }}_{{{m}},{{gmp}}}} = {{\arg\min}}\left\| {{{\Phi }}_{{m}}^{{T}}{{{\Phi }}_{{m}}}{{{\alpha }}_{{m}}} - {\textbf{1}_{{N}}}} \right\|_2^2 + {{\lambda }}\left\| {{{\Phi }}{{{\alpha }}_{{m}}}} \right\|_2^2
\end{equation}
For convenience, we substitute $K$  for $\Phi^T_m \Phi_m$. The analytical solution to the dual formulation is:
\begin{equation}
{{{\alpha }}_{{{m}},{{gmp}}}} = {\left( {K + {{\lambda I}}} \right)^{ - 1}}{\textbf{1}_{{N}}}
\end{equation}
The analytical solution indicates that we can leverage the individual items of $\alpha_m$ which are the weights of the Fisher vector codes in the $m$-th subtasks. If the weight is zero, then the corresponding descriptor makes no contributions in the pooling. In other words, the $m$-th component of the Fisher Vector code is sparsified, whose idea is like FV sparsity encoding in ~\cite{DBLP:journals/ijcv/SanchezPMV13}.

In LLC~\cite{DBLP:conf/cvpr/WangYYLHG10}, weighted L2-norm constraint is used to assure that the local atoms are preserved, which inspires us to use a similar regularity to leverage the sparsity of $\alpha_m$, let $\alpha_{m,{sfv}}$ denote the SFV representation,
\begin{equation}
{{{\alpha }}_{{{m}},{{sfv}}}} = {{\arg\min}}\left\| {{{K}}{{{\alpha }}_{{m}}} - {\textbf{1}_{{N}}}} \right\|_2^2 + {{\lambda }}\left\| {{{d}} \odot {{{\alpha }}_{{m}}}} \right\|_2^2
\end{equation}
where $\emph{d}$ gives different constraints to the individual items of $\alpha_m$. Specially,
\begin{equation}
d_j^m = \left\{ \begin{array}{l}
{\kern 1pt} {\kern 1pt} {\kern 1pt} {\kern 1pt} {\kern 1pt} {\kern 1pt} {\kern 1pt} 1{\kern 1pt} {\kern 1pt} {\kern 1pt} {\kern 1pt} {\kern 1pt} {\kern 1pt} {\kern 1pt} {\kern 1pt} {\kern 1pt} {\kern 1pt} {\kern 1pt} {\kern 1pt} {\kern 1pt} if{\kern 1pt} {\kern 1pt} {\kern 1pt} j{\kern 1pt} {\kern 1pt} {\kern 1pt} {\kern 1pt} {\kern 1pt}  \in {\kern 1pt} {\kern 1pt} {\kern 1pt} {\kern 1pt} \mathcal{N}_\emph{k}^m{\kern 1pt} \\
{\kern 1pt} {\kern 1pt} {\kern 1pt} {\kern 1pt} {\kern 1pt} {\kern 1pt} {\kern 1pt} \infty {\kern 1pt} {\kern 1pt} {\kern 1pt} {\kern 1pt} {\kern 1pt} {\kern 1pt} {\kern 1pt} {\kern 1pt} otherwise
\end{array} \right.{\kern 1pt} {\kern 1pt} {\kern 1pt}
\end{equation}
$\mathcal{N}_\emph{k}^m$ denotes the first $\emph{k}$ maximum posterior of  $\emph{m}$-th cluster.
The analytical solution is:
\begin{equation}
{{{\alpha }}_{{{m}},{{sfv}}}} = {\left( {K^T K + {{\lambda dia}}{{{g}}^2}\left( {{d}} \right)} \right)^{ - 1}}K{\textbf{1}_{{N}}}
\label{eq:kernel}
\end{equation}
When $\lambda$ approaches infinity, $K^T K$
will be comparatively negligible, and the solution can be written as:
\begin{equation}
{{{\alpha }}_{{{m}},{{sfv}}}} = {\left( {{{\lambda dia}}{{{g}}^2}\left( {{d}} \right)} \right)^{ - 1}}K{\textbf{1}_{{N}}}
\label{eq:sparsify}
\end{equation}
Eq. \ref{eq:sparsify} sparsifies the items in $\alpha_m$ that are heavily constrained by \emph{d}, but 
the weights of the unsparsified descriptors are determined by $K{\textbf{1}_{{N}}}$, which is time-costly. Therefore, we make a further simplification.
$K$ is the kernel matrix of patch-to-patch similarities. Clearly $\alpha_{m,sfv}$ only depends on $K$: when a feature shows little similarity with the other features, the corresponding weight $\alpha$ will be greater. Because the Fisher vector codes are all normalized, the diagonal items of $K$ are all ones. If we ignore the non-diagonal items of $K$ which means that the Fisher vector codes are orthogonal, Eq. \ref{eq:sparsify} goes to:
${{{\alpha }}_{{{m}},{{sfv}}}} = {\left( {{{\lambda dia}}{{{g}}^2}\left( {{d}} \right)} \right)^{ - 1}}{\textbf{1}_{{N}}}$.

Because $\lambda$ will be eliminated by normalization, the individual item of $\alpha_{m,sfv}$ can be written as:
\begin{equation}
{{\alpha }}_{{{m}},{{sfv}}}^{\left( {{j}} \right)} =
\left\{
{\begin{array}{*{20}{c}}
{\begin{array}{*{20}{c}}
1&{{{d}}{_j^m} = 1}
\end{array}}\\
{\begin{array}{*{20}{c}}
0&{d{_j^m} = \infty }
\end{array}}
\end{array}} \right.
\label{eq:llc}
\end{equation}
where $\alpha_{m,sfv}^{j}$ is the $\emph{j}$-th term of $\alpha_{m,sfv}$, and $\emph{d}{_j^m}$ is the $\emph{j}$-th term of $\emph{d}{^m}$. As  $\Phi_m\alpha_m$, sparse  $\alpha_m$  makes  $\Phi_m$  be sparsified, i.e., Sparse Fisher vector.

For $\lambda=0$, we have $\alpha_{m,sfv}=\textbf{1}_{{N}} / \lambda$, which corresponds to the original Fisher vector.  Therefore, $\lambda$ does not only play a role in regularization, but also realize a smooth transition between the solution to original Fisher vector ($\lambda=0$) and Sparse Fisher vector ($\lambda\rightarrow\infty$).
\begin{table*}[htbp]
\centering
\caption{Experiment results on PASCAL VOC 2007}
\begin{tabular}{|c|c|c|c|}
\hline
Coding ways & Feature Dims & Accuracy(\%) & Time per image(s) \\
\hline
FV(M=32) & 4096	&49.75 &3.14 \\
FV(M=64) & 8192	&52.48 &5.43 \\
FV(M=128)& 16384 &55.18	&9.73 \\
FV(M=256)& 32768 &58.42	&16.97 \\
\hline
SFV(M=32)&4096   &49.76	&1.04 \\
SFV(M=64)&8192   &52.49	&1.81 \\
SFV(M=128)&16384 &55.18	&2.39 \\
SFV(M=256)&32768	&58.25	&3.72 \\
\hline
BOW(M=8192)&8192 &39.60 &2.45   \\
FV(M=256)[43] &32768 &58.3 &--  \\
\hline
\end{tabular}
\end{table*}

\section{EXPERIMENT EVALUATIONS}
\label{sec:pagestyle}
To verify the effectiveness of Sparse Fisher vector, we validate the proposed approach on image category task.  Firstly, we describe the image classification datasets and experimental setup. We experimentally compare the Sparse Fisher vector against the canonical Fisher vector for two large data sets: Caltech-101 by Fei-Fei et al. ~\cite{DBLP:journals/cviu/Fei-FeiFP07} and the Pascal VOC sets of 2007~\cite{DBLP:journals/ijcv/EveringhamGWWZ10} .
\subsection{Experimental setup}
We compute all SIFT descriptors on overlapping $32\times32$ pixels patches with the step size of 4 pixels. We reduce their dimensionality to 64 dimensions with PCA, so as to better fit the diagonal covariance matrix assumption.

EM algorithm is employed to learn the parameters of the GMM and the cluster number ranges from 64 to 256. By default, for Fisher vector, we calculate the gradient with respect to mean and standard deviation. And for the Sparse Fisher vector we set the neighborhood   as $\emph{k}=5$. We streamline the standard experimental setting and employ linear SVM. It is worth mentioning that the computing platform in our experiments is Intel Core Duo (4G RAM), so the results are slightly different with origin paper in computation time. We use the origin Fisher vector ~\cite{DBLP:conf/cvpr/PerronninD07,DBLP:conf/eccv/PerronninSM10} as the baseline and also the Sparse Fisher vector is improved based on origin Fisher vector.
\subsection{PASCAL VOC 2007}
The Pascal VOC 2007 database contains 9,963 images of 20 classes. We use the standard protocol which consists in training on the provided “trainval” set and testing on the “test” set  and we set the BOW model as the baseline. The classification results are compared in Table 1, where $\emph{M}$ denotes the number of clusters in GMM. We compared three sections in different coding ways, including feature dimensions, accuracy and coding time per image.

For the same feature dimension, for example 8192, the FV achieves higher accuracy than BOW. This result shows that the FV is more discriminative than BOW with the double time cost. But for SFV, when the cluster number of Gaussian mixture distributions(GMM) is 64, we can obtain a comparable accuracy with FV but much faster image coding. This result is in accordance with the conditions of 32, 128 and 256 clusters number.
\subsection{Caltech-101}
Caltech 101 dataset consists of 9144 images of 102 classes like animals, flower and so on. Following the standard experimental setting, we use 30 images per class for training while leaving the remaining for test. Other experimental setting agrees with experiment setup above. Classification results are compared in Table 2.
\begin{table}[htbp]
\centering
\caption{Experiment results on Caltech-101}
\begin{tabular}{|c|c|c|}
\hline
Coding ways & Accuracy(\%) & Time per image(s) \\
\hline
FV(M=32)  &61.00 &1.46 \\
FV(M=64)  &65.09 &2.33 \\
FV(M=128) &67.85	&4.50 \\
FV(M=256) &70.79	&10.69 \\
\hline
SFV(M=32)&61.05	&0.81 \\
SFV(M=64)&65.03	&0.96 \\
SFV(M=128)&67.82	&1.30 \\
SFV(M=256)&70.75	&1.98 \\
\hline
\end{tabular}
\end{table}
Table 2 shows the similar result as Table 1. Under the same size of codebook, SFV runs more quickly than FV with a comparable accuracy. And with the increase in codebook size, the difference of time consuming between these two coding ways is increasing. For example, when the codebook size is 256, coding time per image in FV is 10.69 s, while for SFV is 1.98 s which is nearly 5 times as fast as FV.
\subsection{Experiment analysis}
\label{ssec:subhead}
\subsubsection{Computation cost analysis}
 To further show the advantage of SFV in computation cost, we demonstrate the average coding time per image with the size of codebook and analyze the computation complexity.\\
\begin{figure}[htbp]
\begin{minipage}[t]{0.5\linewidth}
\centering
\includegraphics[width=0.9\textwidth]{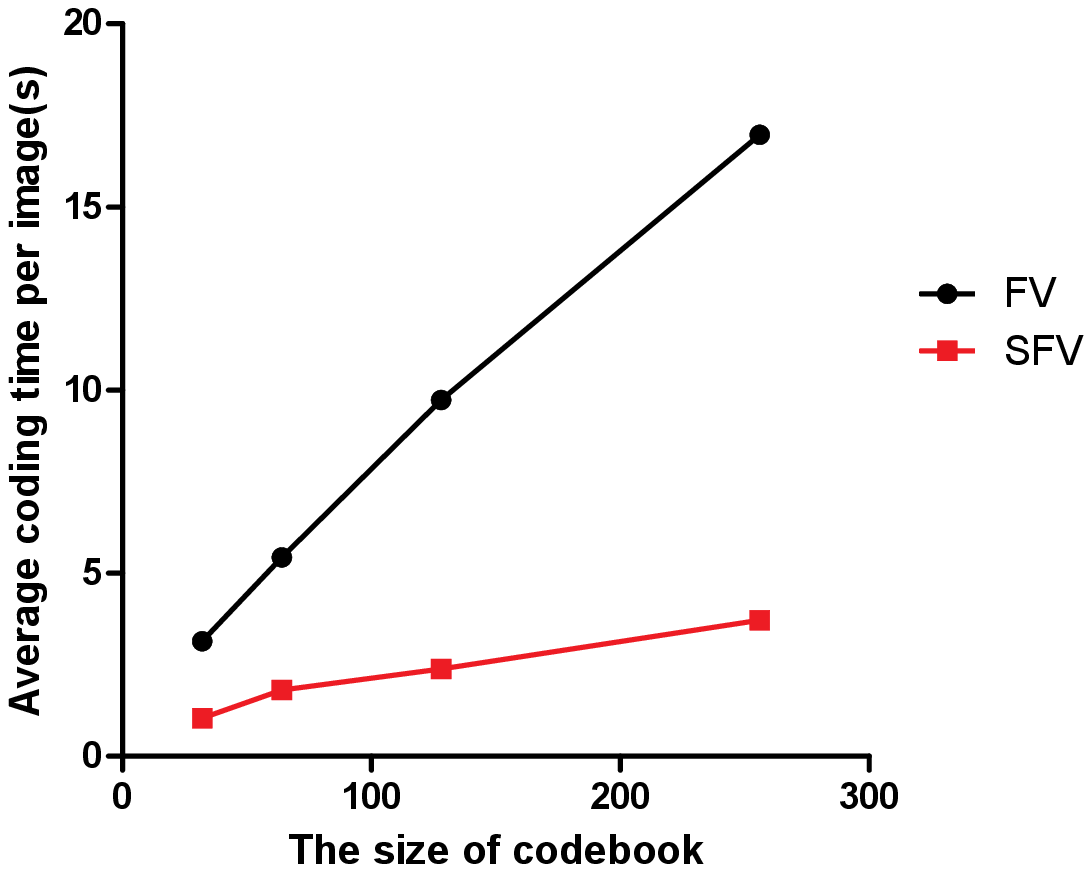}
\subcaption{PASCAL VOC 2007}
\end{minipage}%
\hfill
\begin{minipage}[t]{0.5\linewidth}
\centering
\includegraphics[width=0.9\textwidth]{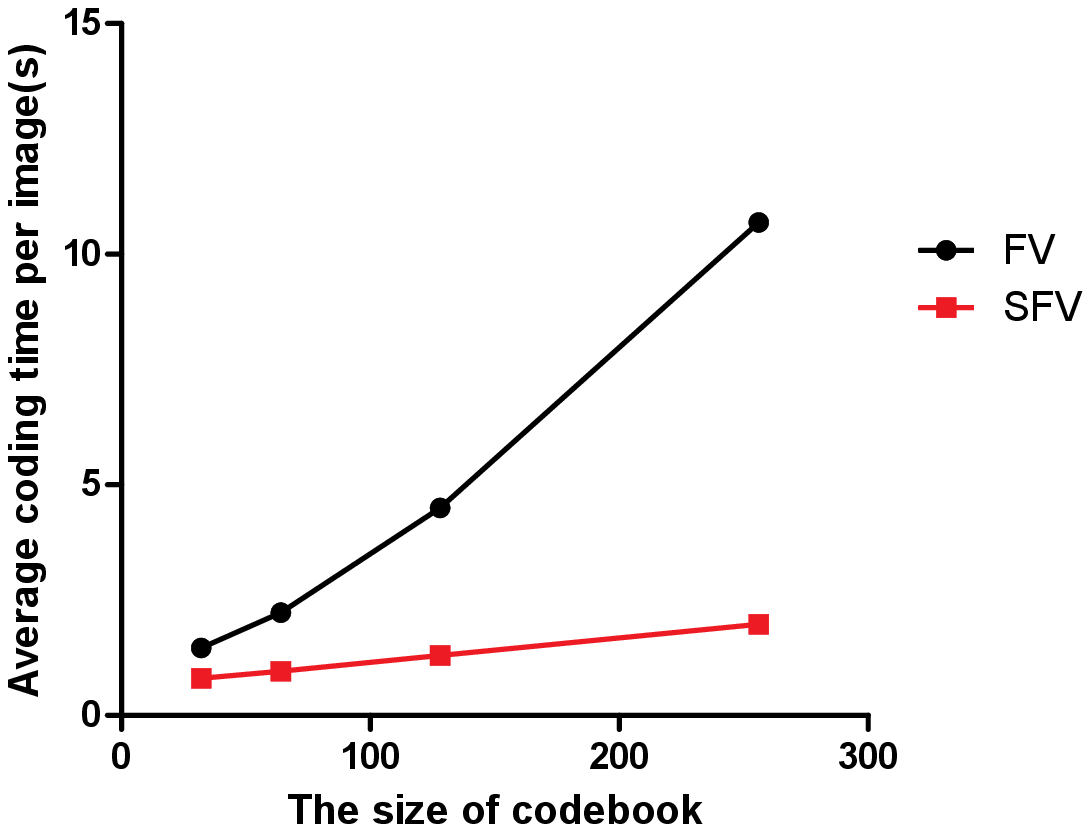}
\subcaption{Caltech 101}
\end{minipage}
\caption{Comparison between theoretical and empirical results, Black line indicate the origin FV and the red one indicate the SFV.}
\end{figure}
%

Fig.1(a) and Fig. 1(b) show the average coding time per image as a function of the codebooks size on datasets above.  As was the case on both datasets, SFV consistently outperforms the FV and the computation time difference increase with the codebooks size.

Considering the $\emph{D}$ dims of features and $\emph{M}$ clusters mentioned above, we can estimate the computation complexity. There are two sub-steps in FV encoding steps: the first sub-step is calculating the posterior probability and the second sub-step is calculating the derivation on the GMM. The computation complexity of the first step is $O[3MD]$ which is same for FV and SFV. The computaion complexity of the second step is $O[(3+5)MD]$ and $O[(3+5)kD]$ respectively. As $M\gg k$ , so the total time of SFV is much less than FV and the time difference increases with $\emph{M}$ which is consist with experiment results.

\subsubsection{Similarity correspondence between SIFT and Sparse Fisher vector}
\label{sssec:subsubhead}
One implicit contribution of our work is that SFV better preserves similarity. To demonstrate this, 200 SIFT features from PASCAL VOC 2007 are randomly selected. We calculate the pair-wise similarity by using cosine measure. The similarity correspondence is shown in Fig. 2. Fig. 2 indicates an obvious linear trend of the similarity between SFV against the similarity between SIFT features, while FV does not. The comparison confirmed that the effectiveness of preserving configuration space locality during coding, which makes similar inputs correspond to similar codes~\cite{DBLP:conf/cvpr/GaoTCZ10,DBLP:journals/pami/GaoTC13}.
\begin{figure}[htbp]
\begin{minipage}[t]{0.48\linewidth}
\centering
\includegraphics[width=1\textwidth]{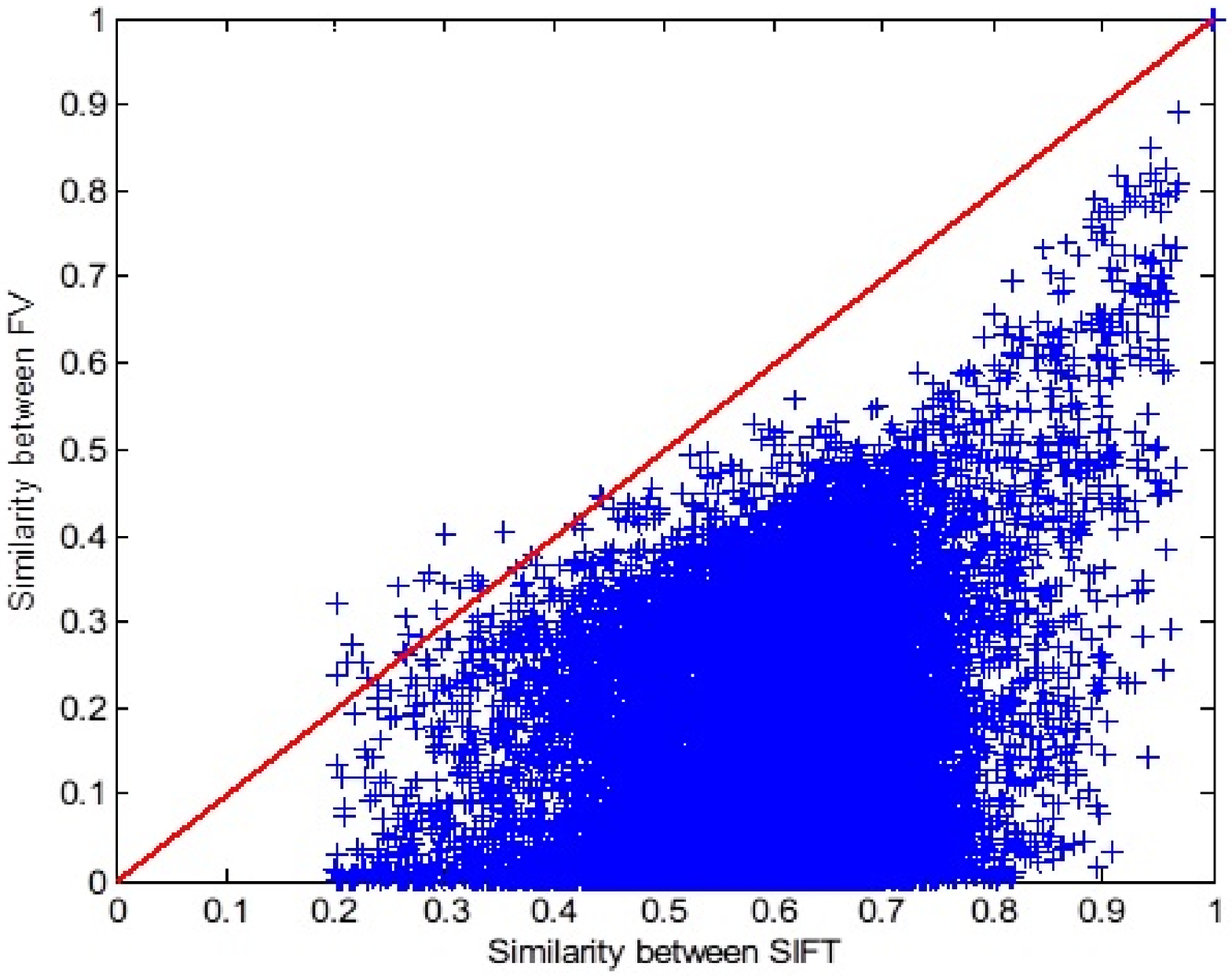}
\subcaption{FV}
\end{minipage}
\hfill
\begin{minipage}[t]{0.48\linewidth}
\centering
\includegraphics[width=1\textwidth]{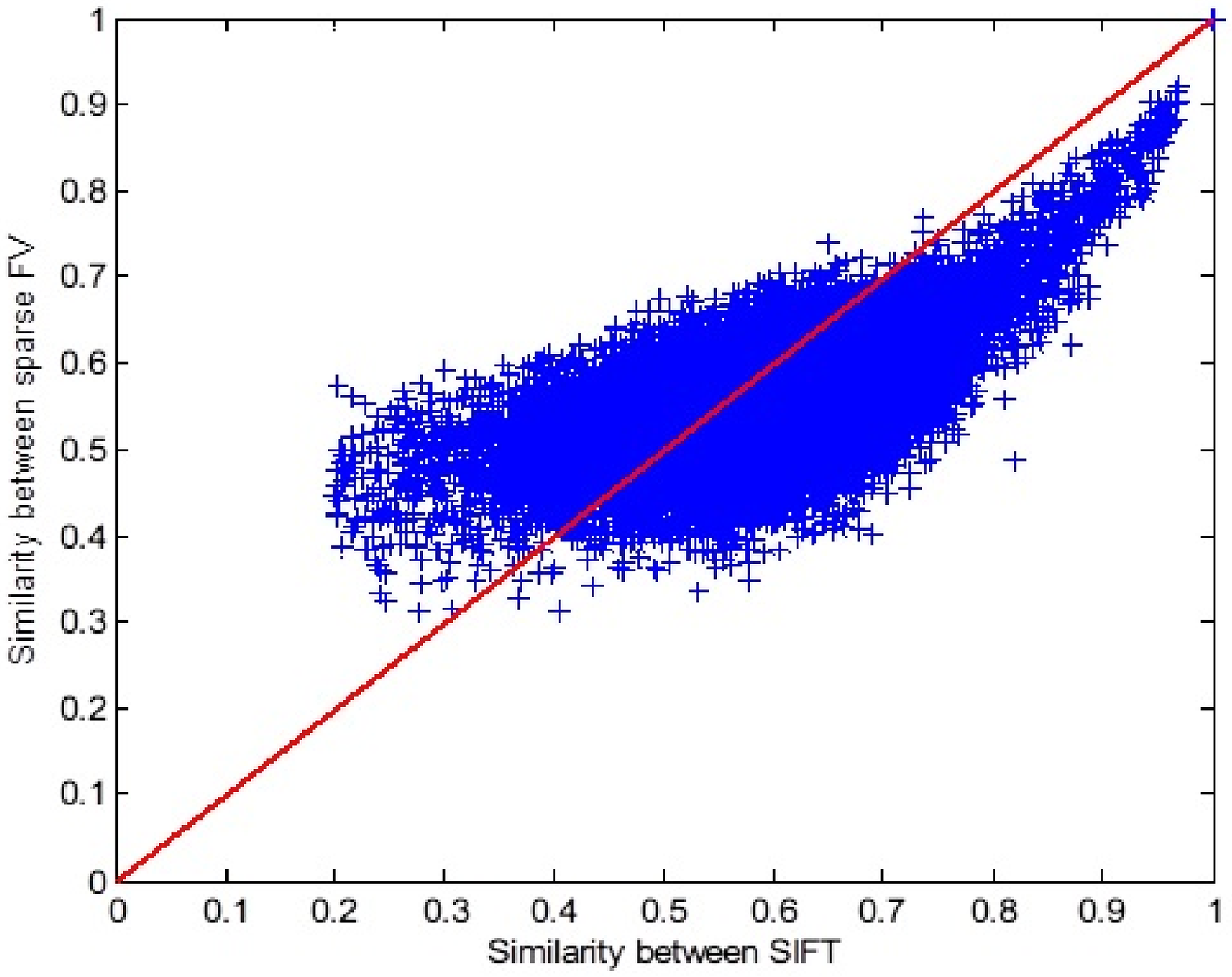}
\subcaption{SFV}
\end{minipage}
\caption{Experiment result on Pascal VOC 2007. The similarity correspondence relationship between the FV(left) or SFV(right) and the SIFT feature. A linear trend can be found in SFV.}
\end{figure}
\subsection{Discussion about SFV}
\label{ssec:subhead}
In Fisher vector, local features are described by deviation from a GMM. The probability representation of a feature by GMM can be represented as:
\begin{equation}
\begin{aligned}
p\left( {x|\theta } \right) &= \sum_{m = 1}^M \omega_m p_m\left( {x|\theta } \right) \\
p_m\left( {x|\theta } \right) &= \frac{exp(-\frac{1}{2}(x-\mu_m)^T \Sigma^{-1}_m (x-\mu_m))}{(2\pi)^{D/2} \left| \Sigma_m \right|^{1/2} }
\label{eq:Fisher}
\end{aligned}
\end{equation}
where  $\omega_{m}$ denotes the prior of the codeword and ${p_m}\left( {x|\theta }\right)$ reflects the probability of feature $\emph{x}$ belongs to the $\emph{m}$-th cluster.
So we can regard the feature coding coefficient as the probability of a feature belonging to the codebook. We notice that no matter in LLC ~\cite{DBLP:conf/cvpr/WangYYLHG10}, or LSC~\cite{DBLP:conf/iccv/LiuWL11}, codewords in codebook are independent and there are no priors on them or we can regard the priors as equal.
For LSC, Eq.\ref{eq:Fisher} can be rewritten as:
\begin{equation}
\begin{aligned}
p\left( {x|B} \right) &= \sum\limits_{m = 1}^M {{p_m}\left( {x|B} \right)} \\
p_m\left( {x|B} \right) &= \exp \left( {\left\| {x - {b_m}} \right\|_2^2/\sigma } \right)
\label{eq:lsc}
\end{aligned}
\end{equation}
Eq.\ref{eq:lsc} can be seen as the probability of input feature $\emph{x}$ belonging to the $\emph{m}$-th codeword ~\cite{DBLP:journals/pami/HuangWWT14}, where $\emph{M }$ denotes the number of codewords in codebook. So the object function of LSC can be represented as:
\begin{equation}
\begin{aligned}
\max    P(x|B) &= \sum\limits_{m = 1}^M {{P_m}(x|B) \odot I(m)} \\
 s.t \quad {\left\| I \right\|_0} &= k
 \label{eq:lsc1}
\end{aligned}
\end{equation}
where $\emph{I}$ is a binary vector.

Also we need to notice that all dimensions of soft coding~\cite{DBLP:conf/iccv/LiuWL11,DBLP:journals/pami/GemertVSG10} are independent of each other. In Fisher coding, the relations among different dimensions are represented by GMM.
The object function of SFV can be represented as:
\begin{equation}
\begin{aligned}
\max    \gamma (m) &= P(m|x,\theta ) = \frac{{P(m){P_m}(x|\theta )}}{{\sum {P(m){P_m}(x|\theta )} }} \odot I(m)\\
 s.t \quad {\left\| I \right\|_0} &= k
\end{aligned}
\end{equation}
where  $\emph{I }$ is a binary vector.

So when we execute the localization operation in Eq. \ref{eq:lsc}, we calculate the codewords which belong to the $\emph{k}$-nearest neighborhood of the feature. This can be regarded as the soft maximum of the likelihood of conditional probability. This is also true for LLC model.
But in SFV, when we execute the early cut off operation, the prior of the codeword is incorporated. So we calculate the codewords which belong to the $\emph{k}$-nearest neighborhood of the feature as Eq. \ref{eq:Fisher}. This can be regarded as a soft maximum of the posterior probability.

\section{Conclusion}
\label{sec:typestyle}
In this paper, we have introduced a 'localized' Fisher vector called Sparse Fisher vector. Based on GMP, we sparsified the Fisher vector code matrix by adding local regular term. These ways allow efficient image categorization without undermining its performance on several public datasets and coding outputs preserve the similarity among input features.

Fisher vector origins from the natural gradient in ~\cite{DBLP:journals/neco/Amari98}, so Sparse Fisher vector can be seen as partial gradient descent. Also, from probabilistic perspective, Sparse Fisher vector can be regarded as a soft maximum of the posterior probability. Since GMP considers the uniqueness of features and weight them according to uniqueness,  we will combine it in our future work.
\section{Acknowledgement}
\label{sec:typestyle}
This work was supported in part by the National Basic Research
Program of China(2012CB719903).
\section{REFERENCES}
\vspace{-2em}
\renewcommand\refname{}
\bibliographystyle{IEEEbib}
\bibliography{referrence2}

\end{document}